\documentclass[conference]{IEEEtran}
\IEEEoverridecommandlockouts
\usepackage{cite}
\usepackage{amsmath,amssymb,amsfonts}
\usepackage{algorithmic}
\usepackage{graphicx}
\usepackage{textcomp}
\usepackage{xcolor}
\usepackage{enumitem}
\usepackage{amsmath}
\usepackage{array}
\newcolumntype{P}[1]{>{\centering\arraybackslash}p{#1}}
\DeclareMathOperator*{\argmin}{arg\,min}
\def\BibTeX{{\rm B\kern-.05em{\sc i\kern-.025em b}\kern-.08em
    T\kern-.1667em\lower.7ex\hbox{E}\kern-.125emX}}
\begin{document}

\title{UNICAD: A Unified Approach for Attack Detection, Noise Reduction and Novel Class Identification 

\thanks{This work is supported by the Engineering and Physical Sciences Research Council [Grant number:EP/V026763/1]}
}

\author{\IEEEauthorblockN{Alvaro Lopez Pellicer, Kittipos Giatgong, Yi Li, Neeraj Suri, Plamen Angelov}
\IEEEauthorblockA{\textit{Lancaster University, UK}\\
\{a.lopezpellicer, giatgong, y.li154, p.angelov, neeraj.suri\}@lancaster.ac.uk}}


\maketitle

\begin{abstract}
As the use of Deep Neural Networks (DNNs) becomes pervasive, their vulnerability to adversarial attacks and limitations in handling unseen classes poses significant challenges. The state-of-the-art offers discrete solutions aimed to tackle individual issues covering specific adversarial attack scenarios, classification or evolving learning. However, real-world systems need to be able to detect and recover from a wide range of adversarial attacks without sacrificing classification accuracy and to flexibly act in {\bf unseen} scenarios. In this paper, UNICAD, is proposed as a novel framework that integrates a variety of techniques to provide an adaptive solution. 

For the targeted image classification, UNICAD achieves accurate image classification, detects unseen classes, and recovers from adversarial attacks using Prototype and Similarity-based DNNs with denoising autoencoders. Our experiments performed on the CIFAR-10 dataset highlight UNICAD's effectiveness in adversarial mitigation and unseen class classification, outperforming traditional models.
\end{abstract}


\section{Introduction}

Machine learning and deep neural networks (DNNs) increasingly enhance efficiency, safety, and decision-making across various sectors \cite{b1, b2}. Although highly efficient in tasks such as image classification \cite{b3, b4, b5}  object detection \cite{b6, b7, b8}, and natural language processing \cite{b9, b10, b11} this technology is exposed to a number of security risks that can be exploited by malicious entities, potentially resulting in harmful outcomes.

Recent research \cite{b12, b13, b14, b15, b17} has shown that deep neural network (DNN) models, crucial for computer vision applications amongst others, are vulnerable to adversarial attacks. These consist on inputs also called adversarial examples specially created to lead models to give incorrect results. Moreover, many machine learning models are inflexible to adapt to new, unseen data categories without the need of retraining and supervision.

Adversarial examples were first introduced by \cite{b12} involve small image alterations that deceive deep learning models. They are, in many cases,  imperceptible to the human eye but can still lead to incorrect model outputs. First examples include Evasion Attack and Box-Constrained L-BFGS \cite{b20}. The transferability of these adversarial examples across different models was later explored in \cite{b13, b15, b17} and underlines further security risks.

Current approaches \cite{b12, b18, b19} have aimed to address challenges concerning adversarial attacks and flexibility on a discrete scenario-by-scenario basis. Systems that are robust to adversarial attacks often do so at the expense of reduced classification accuracy and fail to handle the unknown scenarios. While the opposite happens for systems designed for open set classification.  However in a world where AI is expected to handle such complex tasks such as fully autonomous driving, a framework that provides a unified solution to maintain a robust performance against attacks, standard classification and the unanticipated scenarios is increasingly becoming a necessity.

In this paper, the UNICAD framework (Unified Framework for Adversarial Attack Detection, Noise Reduction, and Novel Class Identification) is proposed as the first integrated approach to these challenges.  UNICAD offers a detection system based on similarity, leveraging DNNs feature extraction capabilities to compare incoming data against known prototypes capable of image classification and identifying any possible adversarial alterations or previously unseen (untrained on) classes leveraging drops in similarity. This is paired with a state-of-the-art denoiser layer, trained to restore data altered by adversarial interference while retaining its essential characteristics without the need for retraining.

The main contributions include:
\begin{itemize}
    \item Novel attack detection supporting robustness to adversarial attacks for prototype and similarity based DNNs explained over Sections III/IV.
    \item Noise reduction via a novel denoising autoencoder schema depicted in Section III-D and Section IV.
    \item Novel class identification through drops in similarity, with the functionality explained over Sections III-E and Section IV.
    \item The unification of state-of-the-art techniques in similarity-based DNNs for attack detection \cite{b37}, interpretable open set classifiers \cite{b42}, and the proposed denoiser autoencoder into a novel architecture, UNICAD. This unified framework is depicted in Section III, with its performance demonstrated in Section IV.
    \item Performance analysis of UNICAD with different backbone networks, including VGG-16 \cite{b49}, DINOv2  \cite{b51} against current state-of-the-art models, in Section IV.
\end{itemize}

\section{Related Work}
Adversarial attacks can be created in different forms, including text \cite{b9, b10, b11}, audio \cite{b21, b22, b23}, and images \cite{b13, b15, b17}. This paper, is focused on image domain attacks, which can be digital (adding adversarial noise to images) or physical (alterations in the real world) \cite{b26, b27, b28}. The most common digital attacks include FGSM \cite{b13}, PGD \cite{b24}, DDN \cite{b25}, and CARLINI \& WAGNER \cite{b17}. 

\subsection{Adversarial Attacks in Image Domain}
For this study, digital attacks in the image domain, specifically FGSM, PGD, and CARLINI \& WAGNER, are the primary focus due to their prevalence and can be described as follows:

\begin{itemize}
    \item Fast Gradient Sign Method (FGSM): FGSM employs the \(L\infty\) metric to create image perturbations as \( \hat{x} = x - \epsilon \cdot \text{sign}(\nabla_x J(\theta, x, y)) \), leveraging \(\nabla_x J\) as the cost function gradient and \(\epsilon\) as a limiting factor \cite{b13}.

    \item Projected Gradient Descent (PGD): Extending FGSM, PGD aims to maximize loss in DNNs, employing \(L_2\) and \(L\infty\) norms. It is formulated as \( \min \rho(\theta), \text{where } \rho(\theta) = E_{(x,y)\sim D}[\max L(\theta, x + \delta, y)] \) \cite{b24}.

    \item Carlini \& Wagner Attack: The approach seeks to find the minimum perturbation size which still produces missclasification. It uses \( \min ||\rho||_p + c \cdot f(x + \rho), \text{s.t. } x + \rho \in [0,1]^m \) \cite{b17}.
\end{itemize}

For illustrative purposes, an example of FGSM attack is provided by Fig. 1 where the attack can be easily seen by the human eye due to the large perturbation size. However effective attacks can easily be crafted to be almost imperceptible.

\begin{figure}[h]
    \centering
    \includegraphics[width=0.45\textwidth]{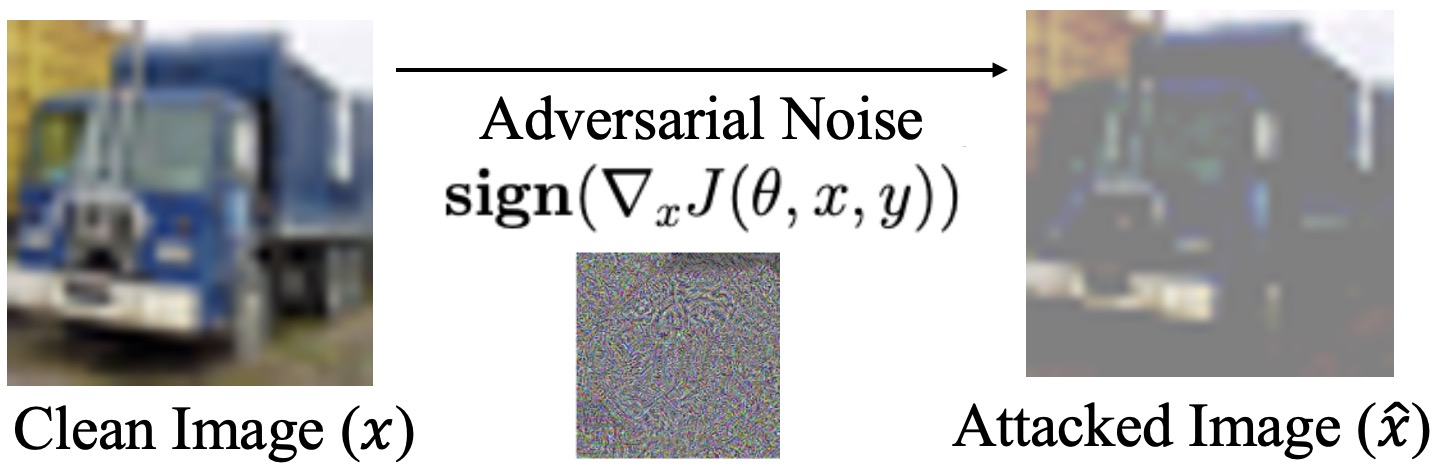}
    \caption{ FGSM attack visualization on a CIFAR-10 image, demonstrating the perturbation effects.}
    \label{fig:mesh1}
\end{figure}

\subsection{Defences Against Adversarial Attacks}
Defences against Adversarial Attacks include a variety of strategies categorized as preventive methods also known as proactive, which includes Adversarial Training \cite{b13}, Input Data Pre-Processing \cite{b18}, Model Ensemble \cite{b29}, Model Regularization \cite{b12}, Model Distillation \cite{b12}, Probable Defences \cite{b30}, Certification and Verification \cite{b31}. And reactive defences such as Detection Mechanisms \cite{b32}, and Denoising/Reconstruction Defences \cite{b33}. Both proactive and reactive defenses often address only specific aspects of adversarial attacks and lack a holistic approach. Moreover Proactive defense often add too much computational overhead. Reactive defences offer a way to detect \cite{b32} or recover attacked images through denoising.

Denoising defences are generally implemented using Denoising Autoencoders (DAEs) \cite{b38, b39}, which are essential in countering adversarial attacks due to their image recovery advantage. However, they are often proposed as a pre-processing step, which adds computational overhead and frequently hinders overall classification accuracy, especially when clean images are processed. This highlights the need for more integrated solutions that can offer a balanced functionality with consistent robustness across various scenarios, without imposing significant computational costs.

\subsection{Novel Class Identification Techniques}
While techniques like xClass \cite{b42}, Deep SVDD \cite{b43}, and the Adaptive Class Augmented Prototype Network \cite{b44} offer effective solutions in their respective areas, they often operate in isolation. This singular focus on specific aspects of the broader challenge limits their practical applicability in dynamic and unpredictable real-world environments. The discrete nature of these solutions underscores the demand for a more holistic approach. A unified framework, such as UNICAD, that seamlessly integrates these different functionalities, could provide a comprehensive solution, ensuring both the reliability and adaptability required for practical applications in diverse settings.

Overall, current state-of-the art offers discrete solutions for image classification, adversarial attack prevention or recovery and novel class classification. However as shown, the non unified use of these techniques bring significant drawbacks that hinders its reliability for practical applications. This calls for unified solutions that can provide a balanced functionality with consistent robustness.

\section{Proposed UNICAD Architecture}

\begin{figure*}[h]
    \centering
    \includegraphics[width=1\textwidth]{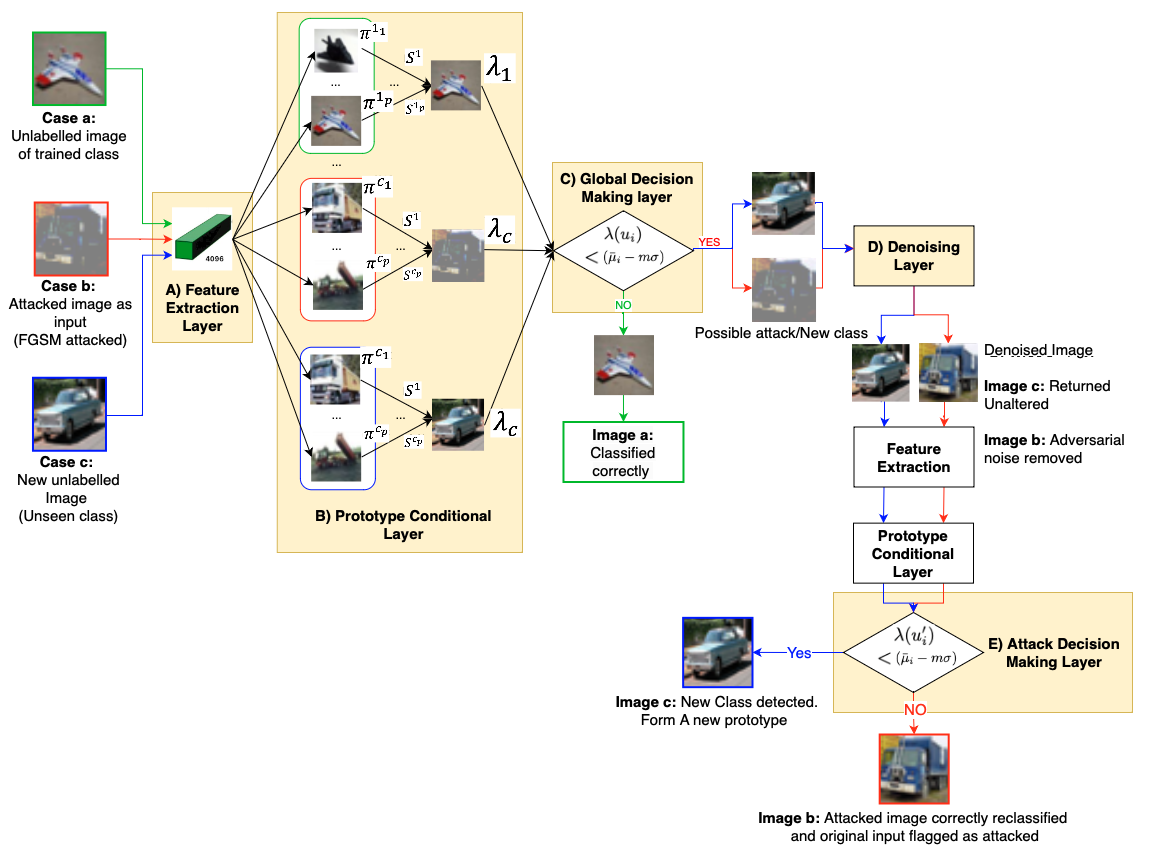}
    \caption{Schematic of the UNICAD framework, showing its layered structure and the processing pathways for (a) trained class images, (b) adversarially attacked images, and (c) unseen class images.}
    \label{fig:mesh1}
\end{figure*}

UNICAD is proposed to address the evolving nature of adversarial threats and limitations in current defenses. UNICAD distinguishes itself by not only assimilating various state-of-the-art techniques but by integrating them into a cohesive, innovative, and efficient novel architecture.
This framework represents a paradigm shift from traditional approaches, maintaining together the strengths of similarity-based attack detection from \cite{b37}, advanced noise reduction via a novel DAE in layer D, and novel class identification inspired by \cite{b42}.

UNICAD's design not only merges but also enhances existing methods with unique elements such as the Denoising Layer (Layer D) and the Attack Decision Layer (Layer E). These new components enable UNICAD to classify images—whether they're clean, non-novel, or not attacked—more efficiently than conventional systems that depend heavily on pre-processing defenses. This approach not only saves computational resources but also makes UNICAD more responsive to adversarial challenges and new class types.

At its heart, UNICAD is built for strong adversarial attack detection, effective noise reduction with denoising auto-encoders, and precise recognition of novel classes using similarity-based neural networks. The framework uses a variation of the Sim-DNN approach \cite{b37} for detecting attacks, employing prototype comparison to spot adversarial intrusions. UNICAD enhances this capability by adding mechanisms to respond to detected attacks and broadening its range to include the identification and handling of unknown classes.

A crucial feature of UNICAD is its ability to detect shifts in data trends or concept drifts. It does this by closely monitoring changes in similarity scores between new data and pre-established prototypes from the training phase. If the similarity score drops below a certain threshold, a possible deviation from the norm is signaled, which could be an adversarial attack or a new class.

Upon detecting such anomalies, UNICAD directs the data to its Denoising Layer, where any adversarial noise is removed. The cleaned images are then reassessed within the framework. If a denoised image is then consistent with an existing class, indicating no further similarity drop, the original image is flagged as attacked but correctly classified, effectively removing the initial adversarial threat. In cases where the similarity drop persists, UNICAD interprets this as an indication of a new class and accordingly creates a new prototype for this unidentified category. This tailored, class-specific approach underlines UNICAD's adaptability, novelty and wide-ranging applicability

The overall architecture of UNICAD is illustrated in Fig. 2 as a layered design comprising of:

\begin{enumerate}[label=\Alph*)]
    \item The Feature Extraction Layer
    \item The Prototype Conditional Probability Layer
    \item The Global Decision Making Layer
    \item The Denoising Layer
    \item The Attack Decision Making Layer
\end{enumerate}

\subsection{Feature Extraction Layer}
UNICAD's feature extraction layer, drawn from \cite{b37}, determines necessary feature space for the algorithm. Due to the adaptable design of UNICAD, features from various sources as network inputs can be incorporated. Feature extraction sources can include CNNs \cite{b45}, Resnets  \cite{b46}, Transformers \cite{b47}, and ensembles \cite{b48}. In this study, two different methods are explored: VGG–16 \cite{b49} due to its high efficiency in this domain \cite{b50} and for direct comparison with related work \cite{b37, b42}, and DINOv2  \cite{b51} due to its outstanding performance in classification tasks and grow in popularity.

The training dataset for UNICAD is denoted as \( x = \{x_1, \ldots, x_N\} \in \mathbb{R}^n \), with corresponding class labels \( y_1, \ldots, y_C \in \{1, \ldots, C\} \). Here, \( N = N_1 + \ldots + N_C \) represents the total number of training samples, \( n \) is the feature count (dimensionality), and \( C \) is the number of classes in the dataset. In this paper, \( n = 4096 \) when VGG-16 is used and \( n = 1536 \) when DINOv2 is used for feature extraction. The most representative samples in the dataset are selected as class-specific prototypes \( \pi \in P \subset X \). These prototypes facilitate a reasoning process based on the similarity (proximity in feature space) between a data sample and a prototype \cite{b16}. In this case, prototypes are identified as the local density peaks \cite{b53}, essentially the most representative samples from the training set. Therefore, only a select few samples from the entire training set are chosen as prototypes, ensuring the system's efficiency and compatibility with a broad range of devices.

Consequently, \( M_j \) represents the total number of prototypes for class \( j \), with \( M_j = |P_j| \) and the total \( M \) being the sum of \( M_j \) across all classes: \( M = \sum_{j=1}^{C} M_j \). In this approach, more than one prototype is considered for each class, so \( M_j > 1 \) for all \( j \). Any new data sample, \( x \in \mathbb{R}^n \), is associated with the nearest prototype from the sets \( P_1, P_2, \ldots, P_C \); collectively \( P = P_1 \cup P_2 \cup \ldots \cup P_C \). The label, \( L \), is assigned as follows:

\begin{equation}
    L(x) = \argmin_{x \in X, \pi \in P} \mathrm{d}(x, \pi).
\end{equation}

\subsection{Prototype Conditional Probability Layer}

This layer is primarily responsible for assessing similarity to established prototypes as well as the formation of prototypes in training and estimating conditional probabilities based on this similarity. It sets the stage for classification by identifying how closely new data aligns with known data patterns drawing from the previously extracted features.

Neurons are characterized by a similarity measure based on data density, denoted as S as outlined in \cite{b54}. This similarity measure is designed to assess the relative or exchangeable closeness of data points within the data space. The measure is articulated through a specific Cauchy equation, as presented in Eq. 2. Theoretically, this similarity metric, which employs either Euclidean or Mahalanobis distance, takes the form of a Cauchy function, as established in the theoretical framework of \cite{b54}. This is quantified by the following Cauchy equation:
\begin{equation}
    S(x) = \frac{1}{1 + \frac{||x-\mu||^2}{||\sigma||^2}}
\end{equation}
where \(x\) represents a data point, with \(\mu\) and \(\sigma\) denoting the global mean and variance, respectively. This step evaluates the proximity of data samples within the feature space, utilizing either Euclidean or Mahalanobis distance as metrics.

An empirical estimation of conditional probabilities is then performed to estimate the probability of the input belonging to existing prototypes based on the distance measure. A multi-modal probability density function is employed where preset assumptions about the data distribution are avoided. The class probability given a data sample, \(x\), is computed as:
\begin{equation}
    p(y|x) = \frac{\sum_{i=1}^{M} N_i S(x)}{\sum_{i=1}^{M} N_i \int_{-\infty}^{\infty} S(x)dx}
\end{equation}
integrating the similarity measure over the entire data space for a refined class probability estimation.

Lastly, prototype association is performed. Class-specific prototypes, distinguished by their peak similarity values, are independently identified. The association of data samples to these prototypes is determined by the equation:
\begin{equation}
    j^* = \argmin_{i=1..N;j=1..M} ||x_i - \pi_j||^2
\end{equation}
A sophisticated assessment of each input is ensured by thus methodology reliant on the proximity of data points to the strategically determined prototypes.

\subsection{Global Decision Making Layer}
This is where the actual classification takes place. Based on the information processed in the Prototype Conditional Probability Layer, this layer makes the final decision on whether an input belongs to an existing class, is an instance of a new class, or is an adversarial attack. This is achieved by employing the recursive mean \( \bar{\mu_i} \) of a parameter \( \lambda \), formulated as follows:
\begin{equation}
    \bar{\mu_i} = \frac{i-1}{i} \bar{\mu_{i}}-1+ \frac{1}{i} \lambda_i \bar{\mu_1} \quad \text{where} \quad \lambda_1 = \bar{\mu_1}. 
\end{equation}
where \( \lambda \) is the degree of similarity between a new data sample and the nearest prototype as described in \cite{b37}.

The  m-$\sigma$ rule is applied to detect potential attacks. This rule can be depicted through an inequality condition:
\begin{equation}
\begin{split}
    \text{IF} \quad & \lambda(u_i) < (\bar{\mu}_i - m\sigma) \\
    \text{THEN} \quad & \text{(u}_i \in \text{ Potential new attack or rew class and} \\
    & \text{u}_i \text{ forwarded to denoising layer) } \\
    \text{ELSE} \quad & \text{(u}_i \in \text{ Classification Label)}
\end{split}
\end{equation}where \( u_i \) is the unlabeled input.

If this condition is met, it suggests that the system has recognized a divergence or a new data concept, distinct from the established data patterns used to generate the prototypes. If not, it indicates no significant change in the data concept, allowing the algorithm to continue with its standard classification process. This mechanism enables UNICAD to adaptively respond to new data and effectively identify potential adversarial attacks or unseen scenarios.

\subsection{Denoising layer}

The Denoising layer includes a novel autoencoder, triggered by eq. 6, to selectively filter out adversarial noise while retaining image integrity. Its unique loss function and broad training regime enable it to handle attacks known and novel effectively, situated within the decision-making framework. Unlike traditional uses as just a preprocessing step \cite{b19}, placing this autoencoder between the global and attack decision layers allows UNICAD to flag and adapt to various attack types, enhancing real-world application and interpretability.

The proposed denoising autoencoder is trained on the original clean images and images attacked using FGSM with \(\epsilon = 0.3\) and \(\epsilon = 0.03\), the choice of FGSM under different configurations has been chosen to improve trasnferability across different untrained types of attacks.

Its loss function, described in eq. (10) is a linear combination of mean squared error (MSE) loss defined as:
\begin{equation}
\mathcal{L}_{\text{MSE}}(y, y') = \frac{1}{M} \sum_{j=1}^{M} (y_j - y'_j)^2
\end{equation}
where \( y \) is the vector of original pixel values, \( y' \) is the vector of reconstructed pixel values by the autoencoder, \( M \) is the total number of pixels in the image, and \( y_j \) and \( y'_j \) are the individual pixel values of the original and reconstructed images, respectively.

The Structural Similarity Index Measure (SSIM) loss defined as:
\begin{equation}
\mathcal{L}_{\text{SSIM}}(x, x') = 1 - \frac{(2\mu_x \mu_{x'} + C_1)(2\sigma_{xx'} + C_2)}{(\mu_x^2 + \mu_{x'}^2 + C_1)(\sigma_x^2 + \sigma_{x'}^2 + C_2)}
\end{equation}
where \( \mu_x \) and \( \mu_{x'} \) are the average values of the original and reconstructed images, \( \sigma_x^2 \) and \( \sigma_{x'}^2 \) are the variances, \( \sigma_{xx'} \) is the covariance, and \( C_1, C_2 \) are constants used to stabilize the division with weak denominator. The SSIM index is a value between -1 and 1, where 1 indicates perfect similarity.

And Feature-based Loss defined as:
\begin{equation}
\mathcal{L}_{\text{Feat}}(f_{\theta}(x), f_{\theta}(x')) = 1 - \frac{\phi(f_{\theta}(x)) \cdot \phi(f_{\theta}(x'))}{\|\phi(f_{\theta}(x))\| \cdot \|\phi(f_{\theta}(x'))\|}
\end{equation}
where \( f_{\theta} \) is the autoencoder with parameters \( \theta \), \( \phi \) represents the feature extractor,  \( x \) is the input image and \( x' \) is the reconstructed image. The loss is calculated as one minus the cosine similarity between the feature representations of \( x \) and \( x' \), promoting alignment in the feature space.

Consequently, the combined loss function of the autoencoder, given a set of training samples,  is defined as:
\begin{equation}
\begin{split}
\mathcal{L}_{\text{comb}}(x, x') = & \, \omega_{\text{MSE}} \cdot \mathcal{L}_{\text{MSE}}(x, x') \\
& + \omega_{\text{SSIM}} \cdot \mathcal{L}_{\text{SSIM}}(x, x') \\
& + \omega_{\text{Feature}} \cdot \mathcal{L}_{\text{Feat}}(\psi(x), \psi(x'))
\end{split}
\end{equation}
where \( \psi(x) \) represents the feature vector obtained from the feature extractor (VGG-16 or DINOv2  in the context of this paper), and \( \omega_{\text{MSE}}, \omega_{\text{SSIM}}, \omega_{\text{Feature}} \) are the weights corresponding to the Mean Squared Error, Structural Similarity Index Measure, and Feature-Based loss components, respectively.

Hence, the training formulation for the autoencoder can be defineed as follows:

\begin{equation}
\theta^* = \arg\min_{\theta} \Bigg[ \frac{1}{N} \sum_{i=1}^{N} \bigg( \mathcal{L}_{\text{comb}}(x_i, x_i') + \mathcal{L}_{\text{comb}}(\hat{x}_i, x_i') \bigg) \Bigg]
\end{equation}
where $\theta^*$ represents the optimal parameters for the autoencoder. $\hat{x}_i$ indicates the $i$-th adversarial image. $\mathcal{L}_{\text{comb}}$ is the combined loss function. $N$ is the total number of training samples.

This training approach is crucial for enhancing the model's capability to generalize across both clean and adversarially perturbed inputs. By simultaneously optimizing for clean and attacked images, the autoencoder learns a more robust representation, improving its efficacy in real-world scenarios where input integrity cannot be guaranteed. This methodology ensures not only the fidelity of image reconstruction. It allows for images of unknown classes to be returned unedited both visually and in their latent space representation which is crucial for them to be classified as new class in the attack decision making layer and for attacked images to be reclassified correctly, therefore neutralizing the attack.

The autoencoder's architecture depicted in Fig. 3 is meticulously crafted, featuring an encoder that incrementally compresses the input using a sequence of convolutions, batch normalization, and pooling layers, interspaced with residual blocks. These blocks are instrumental in retaining information across the network's depth. The decoder is tasked with upscaling the condensed representation to reconstruct the original image. Through a series of transposed convolutions and the application of non-linear activation, such as ReLU and Sigmoid, the decoder learns to reverse the encoding effectively.

\begin{figure}[h]
    \centering
    \includegraphics[width=0.49\textwidth]{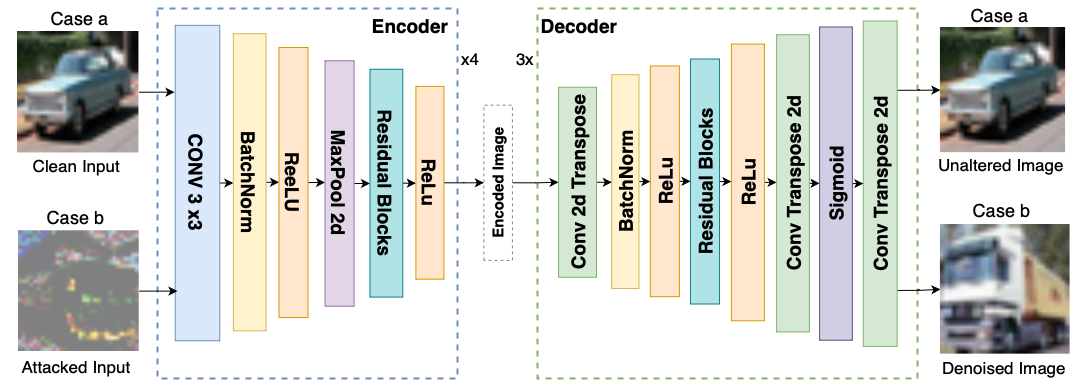}
    \caption{Architecture of the proposed denoising autoencoder highlighting two scenarios: (a) processing of clean images and (b) removal of adversarial noise from attacked images.}
    \label{fig:mesh1}
\end{figure}

This autoencoder has shown state-of-the-art results which have been displayed in Section IV. A visual example of its performance both in attacked images and clean images is also shown in Fig. 3.

\subsection{Attack Decision Making Layer}
Reconstructed images are forwarded to this last layer. Here the reconstructed input is fed forward into the framework once more. After the reconstructed input is processed by layers A, and B, a decision about the suspected input is taken as follows: 
\begin{equation}
\begin{split}
    \text{IF} \quad & \lambda(u_i') < (\bar{\mu}_i - m\sigma) \\
    \text{THEN} \quad & \text{(u}_i \in \text{ New class.} \\
    & \text{Form a new prototype)}  \\
    \text{ELSE} \quad & \text{(u}_i \in \text{Correct Classification Label and} \\
    & \text{original input flagged as attack)}
\end{split}
\end{equation}
where \( u_i' \) the reconstructed input.

It is important to note that inputs flagged as attacked could in the future be used as prototypes of adversarial attacks to develop evolving attack detectors and attack classifiers with visual representations for human interpretability.

\section{Results and discussion}
\subsection{Dataset}
A variety of attack scenarios have been performed to validate the robustness of the UNICAD framework. Experiments have been performed using the standard CIFAR-10 \cite{b52} datasets given their popularity as a benchmark and their complexity in matching real-world situations. For training, the CIFAR-10 training dataset was used, which has 50000 training images, 5000 images per class. The overall training formulation has been performed as follows: 
\subsection{Experimental Configuration} 

For layer, A, when VGG-16 has been chosen as Feature Extractor, a VGG-16 pretrained on ImageNet has been Fine-Tuned for 10 epochs into CIFAR-10, and last FC layer has been used as Feature Extractor. When DINOv2 was used, no Fine-Tuning was performed and the last FC was chosen as FE.

The proposed DAE in layer D was trained for 20 epochs as described in Section III-D. The feature extractor chosen for layer A was chosen for Feature-Based loss in the autoencoder (eq. 9). 

Layer B was trained to find prototypes with CIFAR-10, taking an average of 0.17\% of the images to form the necessary prototypes and an average training time of 27.09 seconds. Layers C and E do not require training. This highlights the training efficiency and adaptability of the core framework.

The CIFAR-10 validation dataset was later used for validation which cointains 10000 different images, 1000 images per class.

The experimental setup for unseen class detection is as follows: UNICAD and methods used for comparison were trained in CIFAR-10 classes 0-8. Hence class 9 was left unseen. The training steps are the same as described in the previous paragraph, but with one class dropped. This means, 45000 images were used in training, 5000 images per class still. For validation, all images from class 9 were used (6000). Although this experiment setup apparently does not account for scenarios where there are multitude of different inputs, as detailed evaluation is done discretely per every input, which means in evaluation  and real-world settings the same resutls would extrapolate.

It is important to note that class 9 correspond to truck images, and class 0 are automobile, so some error is expected as prototypes from truck images and vehicle images are close both visually and on the latent space. Further investigation on the similarity of different classes in the latent space and its relevance is suggested as future work.

The metric considered to evaluate unseen class detection rate is as follows:

\begin{equation}
\text{Detection}(\%) = \frac{TP + TN}{TP + FP + TN + FN} \times 100,
\end{equation}
\subsection{Results}
The ablation study is out of scope of this paper. However, to confirm the effectiveness of each contribution, Table I, shows the difference in capabilities between UNICAD, and the main competitor models: XClass \cite{b42}, SimDNN \cite{b37} and DDSA \cite{b19}  

\begin{table}[h]
\centering
\caption{Comparative overview of capabilities across different methods, including unseen class detection, attack detection, and attack recovery features.}
\label{tab:my_label}
\begin{tabular}{|l|P{1.7cm}|P{1cm}|P{1cm}|}
\hline
\textbf{Method} & \textbf{Unseen Class \newline Detection} & \textbf{Attack \newline Detection} & \textbf{Attack \newline Recovery} \\ \hline
XClass \cite{b42}         & $\checkmark$                   & $\times$                  & $\times$                 \\ \hline
simDNN \cite{b37}          & $\times$                       & $\checkmark$              & $\times$                 \\ \hline
DDSA \cite{b19}            & $\times$                       & $\times$                  & $\checkmark$             \\ 
\hline
\textbf{UNICAD (ours)}            & $\checkmark$                   & $\checkmark$              & $\checkmark$             \\ 
\hline
\end{tabular}
\end{table}

\begin{table*}[h!]
\centering
\caption{ Performance comparison of mainstream image classifiers and denoisers versus our proposed denoising autoencoder (DAE) across various attack scenarios using CIFAR-10 data.}
\label{table:Denoising}
\begin{tabular}{|l|c|P{1.2cm}|P{1.2cm}|P{1.2cm}|P{1.2cm}|P{1.2cm}|P{1.2cm}|}
\hline
Model & Clean & PGD  \newline $\boldsymbol{\varepsilon = 0.01}$ & PGD \newline $\boldsymbol{\varepsilon = 0.3}$ & FGSM \newline $\boldsymbol{\varepsilon = 0.01}$ & FGM \newline $\boldsymbol{\varepsilon = 0.03}$ & FGM \newline $\boldsymbol{\varepsilon = 0.3}$ & C\&W \newline $L_2$ norm \\ \hline
Ours (VGG-16 for Feat loss) & 93.10 & 83.58 & 85.82 & 82.67 & 78.10 & 83.26 & 83.2 \\ \hline
Ours (VGG19 for Feat loss) & 93.10 & 83.58 & 85.82 & 82.67 & 78.10 & 83.26 & 83.4 \\ \hline
Ours (DINOv2 for Feat loss) & 97.24 & \textbf{90.55} & \textbf{92.71} & \textbf{90.3} & \textbf{88.87} & \textbf{92.24} & \textbf{93.7} \\ \hline
DDSA \cite{b19} & 81.4 & 60.00 & 61.10 & 63.00 & 61.00 & 57.10 & 36.70 \\ \hline
VGG-16 No defence & 92.0 & 0.05 & 32.12 & 31.06 & 22.56 & 0.11 & 0.00 \\ \hline
DINOv2  No defence & \textbf{97.63} & 56.8 & 0.7 & 58.9 & 17.3 & 12.4 & 0.8 \\ \hline
\end{tabular}
\end{table*}

\begin{table*}[h!]
\centering
\caption{ UNICAD framework compared to mainstream models under multiple scenarios, including clean images, different adversarial attacks, and unseen class detection using CIFAR-10 data.}
\label{table:UNICAD}
\begin{tabular}{|l|P{1.6cm}|P{1.6cm}|P{1.6cm}|P{1cm}|P{1.5cm}|P{1.5cm}|}
\hline
Scenario & UNICAD \newline  (VGG-16 FE) & UNICAD \newline (DINOv2  FE) & xClass \newline  (VGG-16 FE) & DDSA \newline \cite{b19} & VGG-16 \newline No defence & DINOv2  \newline No defence \\ \hline
Clean & 80.86 & 92.93 & 80.86 & 81.4 & 92.0 & \textbf{97.63} \\ \hline
PGD $\boldsymbol{\varepsilon = 0.01}$ & 74.81 & \textbf{77.77} & 49.3 & 60.00 & 0.05 & 56.8 \\ \hline
PGD $\boldsymbol{\varepsilon = 0.3}$ & 72.63 & \textbf{82.29} & 14.2 & 61.10 & 32.12 & 0.7 \\ \hline
FGSM $\boldsymbol{\varepsilon = 0.01}$ & 70.01 & \textbf{77.37} & 49.0 & 63.00 & 31.06 & 58.9 \\ \hline
FGM $\boldsymbol{\varepsilon = 0.03}$ & 64.9 & \textbf{76.02} & 47.1 & 61.00 & 22.56 & 17.3 \\ \hline
FGM $\boldsymbol{\varepsilon = 0.3}$ & 73.09 & \textbf{81.10} & 15.6 & 57.10 & 0.11 & 12.4 \\ \hline
C\&W $L_2$ norm & 73.2 & \textbf{79.33} & 0.6 & 36.70 & 0.00 & 0.8 \\ \hline
Unseen Class detection & 62.30 & \textbf{83.38} & 62.30 & 0.00 & 0.00 & 0.00 \\ \hline
\end{tabular}
\end{table*}

Table II shows the performance of the denoising autoencoder part of the denoising layer using VGG-16 and DINOv2  as the backbone for Feature-based loss (eq. 9) in a clean image setting and against FGSM, PGD and C\&W attacks at different settings. The performance of VGG-16 and DINOv2  under this attacks with no defence has been provided along with results of current state-of-the-art DAEs, namely DDSA \cite{b19}. Accuracy has been evaluated on CIFAR-10 testing dataset for each scenario. Accuracy values displayed are measured in percentage, and it is measured from classification of the reconstructed input images in each scenario by the chosen classifiers, VGG-16 or DINOv2, depending on the feature backbone of the network. Results demonstrate that the proposed denoiser autoencoder works better than current state of the art, with DINOv2  performing best as the backbone for Feature-based loss as expected due to its standalone performance. It is also very remarkable that Clean accuracy after reconstruction drops by less than 1\% in all cases compared to current state-of-the-art denoisers where that drop is much more pronounced. This is due to the novel combined loss function that preserves both latent space and visual integrity of images. Robustness against no defence is also stated with consistently over 80\% results in the defence approach compared to the performance of classifiers where no defence is implemented with attacks managing to drop their accuracy to as low as 0\% in the worst case scenarios.

Table III shows the performance of the overall framework with VGG-16 and DINOv2  as the backbone feature extractors for the Feature extraction layer. In this table classification accuracy of attacked images using FGSM, PGD and C\&W have been included as well as clean image classification and unseen class classification along with resutls of current state-of-the-art DAEs, namely DDSA \cite{b19}, and xClass \cite{b42}. Accuracy is presented in percentage and accuracy for the UNICAD framework has been calculated at the global decision making layer for clean images and at the attack decision making layer for attacked and unseen images.

Accuracy is notably lower than that of the standalone denoising autoencoder, attributed to the new configuration's ability to detect unseen classes and provide interpretability via prototype-based classification. Despite this, the results remain robust, with DINOv2's classification accuracy dropping  $<$ 5\% compared to its undefended setup. In clean accuracy, the performance surpasses DDSA when using VGG-16, with DINOv2 yielding the best outcomes, particularly in clean accuracy, thanks to its superior feature extraction capabilities.

xClass \cite{b42} has been included for comparison due to its unseen class detection capability, despite not being robust against adversarial attacks. It is matched by UNICAD in classification and new class identification. UNICAD excels with consistant $>$ 70\% accuracy in the presence of adversarial attacks, outperforming mainstream methods and affirming its theoretical robustness against such attacks and effectiveness in detecting new classes.

\section{Conclusion and Future Work}

Overall, UNICAD, a novel framework designed to increase the robustness of deep neural networks against adversarial attacks and enable the detection of novel, unseen classes has been presented. Different experiments using CIFAR-10 have been shown which demonstrated not only UNICAD's capability to withstand various adversarial attacks but also its ability to identify and classify new, unseen classes with high accuracy. The integration of a novel state-of-the-art denoising autoencoder within the denoising layer significantly increased the framework's robustness to adversarial attacks. Furthermore the limitations of the current state-of-the-art and the impending need for a unified framework has been demonstrated throughout the paper.

Another key element of UNICAD is its prototype-based architecture, which has contributed to both the interpretability of the system and its ability to adapt to new data scenarios without extensive retraining. This feature is especially crucial in dynamic environments where models frequently encounter data outside their initial training distribution such as Autonomous Systems.

While UNICAD has marked a significant advancement in defensive strategies against adversarial attacks and unseen class detection, the ongoing arms race in adversarial machine learning is acknowledged. Future work will be aimed to further enhance the scalability and efficiency of UNICAD, particularly focusing on optimizing the denoising layer to provide better adaptability in a landscape of changing and evolving adversarial attacks and performing experimentation across a wider settings of datasets to further showcase UNICAD's performance along with robust ablation studies, which are acknowledged as a limitation of this paper, which due to page limitations make such experiments out of the scope for this paper.

\bibliographystyle{IEEEtran}
\bibliography{UNICAD}

\begin{thebibliography}{10}
\providecommand{\url}[1]{#1}
\csname url@samestyle\endcsname
\providecommand{\newblock}{\relax}
\providecommand{\bibinfo}[2]{#2}
\providecommand{\BIBentrySTDinterwordspacing}{\spaceskip=0pt\relax}
\providecommand{\BIBentryALTinterwordstretchfactor}{4}
\providecommand{\BIBentryALTinterwordspacing}{\spaceskip=\fontdimen2\font plus
\BIBentryALTinterwordstretchfactor\fontdimen3\font minus \fontdimen4\font\relax}
\providecommand{\BIBforeignlanguage}[2]{{%
\expandafter\ifx\csname l@#1\endcsname\relax
\typeout{** WARNING: IEEEtran.bst: No hyphenation pattern has been}%
\typeout{** loaded for the language `#1'. Using the pattern for}%
\typeout{** the default language instead.}%
\else
\language=\csname l@#1\endcsname
\fi
#2}}
\providecommand{\BIBdecl}{\relax}
\BIBdecl

\bibitem{b1}
T.~B. Sheridan, ``Human–robot interaction: status and challenges,'' \emph{Human Factors}, vol.~58, no.~4, pp. 525--532, 2016.

\bibitem{b2}
B.~Siciliano and O.~Khatib, Eds., \emph{Springer handbook of robotics}, 2nd~ed.\hskip 1em plus 0.5em minus 0.4em\relax Springer Cham, jul 2016.

\bibitem{b3}
X.~Chen, C.~Liang, D.~Huang, E.~Real, K.~Wang, Y.~Liu, H.~Pham, X.~Dong, T.~Luong, C.-J. Hsieh, Y.~Lu, and Q.~V. Le, ``Symbolic discovery of optimization algorithms,'' \emph{arXiv preprint arXiv: 2302.06675}, 2023.

\bibitem{b4}
K.~Simonyan and A.~Zisserman, ``Very deep convolutional networks for large-scale image recognition,'' \emph{arXiv preprint arXiv: 1409.1556}, 2014.

\bibitem{b5}
A.~Krizhevsky, I.~Sutskever, and G.~E. Hinton, ``Imagenet classification with deep convolutional neural networks,'' \emph{Commun. ACM}, vol.~60, no.~6, p. 84–90, may 2017.

\bibitem{b6}
N.~Dalal and B.~Triggs, ``Histograms of oriented gradients for human detection,'' in \emph{2005 IEEE Computer Society Conference on Computer Vision and Pattern Recognition (CVPR'05)}, vol.~1, 2005, pp. 886--893 vol. 1.

\bibitem{b7}
J.~Redmon, S.~Divvala, R.~Girshick, and A.~Farhadi, ``You only look once: Unified, real-time object detection,'' \emph{arXiv preprint arXiv: 1506.02640}, 2016.

\bibitem{b8}
R.~Girshick, ``Fast r-cnn,'' \emph{arXiv preprint arXiv: 1504.08083}, 2015.

\bibitem{b9}
M.~Alzantot, Y.~Sharma, A.~Elgohary, B.-J. Ho, M.~Srivastava, and K.-W. Chang, ``Generating natural language adversarial examples,'' \emph{arXiv preprint arXiv: 1804.07998}, 2018.

\bibitem{b10}
J.~Ebrahimi, A.~Rao, D.~Lowd, and D.~Dou, ``Hotflip: White-box adversarial examples for text classification,'' \emph{arXiv preprint arXiv: 1712.06751}, 2018.

\bibitem{b11}
M.~Iyyer, J.~Wieting, K.~Gimpel, and L.~Zettlemoyer, ``Adversarial example generation with syntactically controlled paraphrase networks,'' \emph{arXiv preprint arXiv: 1804.06059}, 2018.

\bibitem{b12}
C.~Szegedy, W.~Zaremba, I.~Sutskever, J.~Bruna, D.~Erhan, I.~Goodfellow, and R.~Fergus, ``Intriguing properties of neural networks,'' \emph{arXiv preprint arXiv: 1312.6199}, 2013.

\bibitem{b13}
I.~J. Goodfellow, J.~Shlens, and C.~Szegedy, ``Explaining and harnessing adversarial examples,'' \emph{arXiv preprint arXiv: 1412.6572}, 2014.

\bibitem{b14}
N.~Papernot, P.~McDaniel, S.~Jha, M.~Fredrikson, Z.~B. Celik, and A.~Swami, ``The limitations of deep learning in adversarial settings,'' in \emph{2016 IEEE European Symposium on Security and Privacy (EuroS`I\&'P)}, 2016, pp. 372--387.

\bibitem{b15}
S.-M. Moosavi-Dezfooli, A.~Fawzi, and P.~Frossard, ``Deepfool: A simple and accurate method to fool deep neural networks,'' 2016, pp. 2574--2582.

\bibitem{b17}
N.~Carlini and D.~Wagner, ``Towards evaluating the robustness of neural networks,'' pp. 39--57, 2017.

\bibitem{b20}
R.~Fletcher, \emph{Practical Methods of Optimization}, 2nd~ed., .~John Wiley `I\&'~Sons, Ed.\hskip 1em plus 0.5em minus 0.4em\relax Hoboken, NJ, USA: Wiley, 2013.

\bibitem{b18}
C.~Xie, J.~Wang, Z.~Zhang, Y.~Zhou, L.~Xie, and A.~Yuille, ``Adversarial examples for semantic segmentation and object detection,'' in \emph{2017 IEEE International Conference on Computer Vision (ICCV)}, 2017, pp. 1378--1387.

\bibitem{b19}
Y.~Bakhti, S.~A. Fezza, W.~Hamidouche, and O.~Déforges, ``Ddsa: A defense against adversarial attacks using deep denoising sparse autoencoder,'' \emph{IEEE Access}, vol.~7, pp. 160\,397--160\,407, 2019.

\bibitem{b37}
E.~Soares, P.~Angelov, and N.~Suri, ``Similarity-based deep neural network to detect imperceptible adversarial attacks,'' in \emph{2022 IEEE Symposium Series on Computational Intelligence (SSCI)}, 2022, pp. 1028--1035.

\bibitem{b42}
P.~Angelov and E.~Soares, ``Detecting and learning from unknown by extremely weak supervision: exploratory classifier (xclass),'' \emph{Neural Comput `I\&' Applic}, vol.~33, pp. 15\,145--15\,157, 2021.

\bibitem{b49}
K.~Simonyan and A.~Zisserman, ``Very deep convolutional networks for large-scale image recognition,'' \emph{arXiv preprint arXiv:1409.1556}, 2014.

\bibitem{b51}
M.~Oquab, T.~Darcet, T.~Moutakanni, H.~Vo, M.~Szafraniec, V.~Khalidov, P.~Fernandez, D.~Haziza, F.~Massa, A.~El-Nouby, M.~Assran, N.~Ballas, W.~Galuba, R.~Howes, P.-Y. Huang, S.-W. Li, I.~Misra, M.~Rabbat, V.~Sharma, G.~Synnaeve, H.~Xu, H.~Jegou, J.~Mairal, P.~Labatut, A.~Joulin, and P.~Bojanowski, ``Dinov2: learning robust visual features without supervision,'' \emph{arXiv preprint arXiv: 2304.07193}, 2023.

\bibitem{b21}
N.~Carlini and D.~Wagner, ``Audio adversarial examples: Targeted attacks on speech-to-text,'' in \emph{2018 IEEE Security and Privacy Workshops (SPW)}, 2018, pp. 1--7.

\bibitem{b22}
X.~Yuan, Y.~Chen, Y.~Zhao, Y.~Long, X.~Liu, K.~Chen, S.~Zhang, H.~Huang, X.~Wang, and C.~A. Gunter, ``Commandersong: A systematic approach for practical adversarial voice recognition,'' \emph{arXiv preprint arXiv: 1801.08535}, 2018.

\bibitem{b23}
L.~Schönherr, K.~Kohls, S.~Zeiler, T.~Holz, and D.~Kolossa, ``Adversarial attacks against automatic speech recognition systems via psychoacoustic hiding,'' \emph{arXiv preprint arXiv: 1808.05665}, 2018.

\bibitem{b26}
A.~Kurakin, I.~Goodfellow, and S.~Bengio, ``Adversarial examples in the physical world,'' \emph{arXiv preprint arXiv: 1607.02533}, 2016.

\bibitem{b27}
A.~Athalye, L.~Engstrom, A.~Ilyas, and K.~Kwok, ``Synthesizing robust adversarial example,'' \emph{arXiv preprint arXiv: 1707.07397}, 2017.

\bibitem{b28}
M.~Sharif, S.~Bhagavatula, L.~Bauer, and M.~K. Reiter, ``Accessorize to a crime: Real and stealthy attacks on state-of-the-art face recognition,'' in \emph{Proceedings of the 2016 ACM SIGSAC Conference on Computer and Communications Security}, ser. CCS '16.\hskip 1em plus 0.5em minus 0.4em\relax New York, NY, USA: Association for Computing Machinery, 2016, p. 1528–1540.

\bibitem{b24}
A.~Madry, A.~Makelov, L.~Schmidt, D.~Tsipras, and A.~Vladu, ``Towards deep learning models resistant to adversarial attacks,'' \emph{arXiv preprint arXiv: 1706.06083}, 2017.

\bibitem{b25}
J.~Rony, L.~G. Hafemann, L.~S. Oliveira, I.~B. Ayed, R.~Sabourin, and E.~Granger, ``Decoupling direction and norm for efficient gradient-based l2 adversarial attacks and defenses,'' in \emph{2019 IEEE/CVF Conference on Computer Vision and Pattern Recognition (CVPR)}, 2019, pp. 4317--4325.

\bibitem{b29}
L.~I. Kuncheva and C.~J. Whitaker, ``Measures of diversity in classifier ensembles and their relationship with the ensemble accuracy,'' \emph{Mach. Learn.}, vol.~51, no.~2, p. 181–207, may 2003.

\bibitem{b30}
R.~Ehlers, ``Formal verification of piece-wise linear feed-forward neural networks,'' \emph{arXiv preprint arXiv: 1705.01320}, 2017.

\bibitem{b31}
S.~Gowal, K.~Dvijotham, R.~Stanforth, R.~Bunel, C.~Qin, J.~Uesato, R.~Arandjelovic, T.~Mann, and P.~Kohli, ``On the effectiveness of interval bound propagation for training verifiably robust models,'' \emph{arXiv preprint arXiv: 1810.12715}, 2018.

\bibitem{b32}
N.~Papernot, P.~McDaniel, I.~Goodfellow, S.~Jha, Z.~B. Celik, and A.~Swami, ``Practical black-box attacks against machine learning,'' in \emph{Proceedings of the 2017 ACM on Asia Conference on Computer and Communications Security}, ser. ASIA CCS '17.\hskip 1em plus 0.5em minus 0.4em\relax New York, NY, USA: Association for Computing Machinery, 2017, p. 506–519.

\bibitem{b33}
A.~Buades, B.~Coll, and J.-M. Morel, ``A non-local algorithm for image denoising,'' in \emph{2005 IEEE Computer Society Conference on Computer Vision and Pattern Recognition (CVPR'05)}, vol.~2, 2005, pp. 60--65 vol. 2.

\bibitem{b38}
D.~Meng and H.~Chen, ``Magnet: A two-pronged defense against adversarial examples,'' in \emph{Proceedings of the 2017 ACM SIGSAC Conference on Computer and Communications Security}, ser. CCS '17.\hskip 1em plus 0.5em minus 0.4em\relax New York, NY, USA: Association for Computing Machinery, 2017, p. 135–147.

\bibitem{b39}
P.~Vincent, H.~Larochelle, Y.~Bengio, and P.-A. Manzagol, ``Extracting and composing robust features with denoising autoencoders,'' in \emph{Proceedings of the 25th International Conference on Machine Learning}, ser. ICML '08.\hskip 1em plus 0.5em minus 0.4em\relax New York, NY, USA: Association for Computing Machinery, 2008, p. 1096–1103.

\bibitem{b43}
L.~Ruff, R.~Vandermeulen, N.~Goernitz, L.~Deecke, S.~A. Siddiqui, A.~Binder, E.~M{\"u}ller, and M.~Kloft, ``Deep one-class classification,'' in \emph{Proceedings of the 35th International Conference on Machine Learning}, ser. Proceedings of Machine Learning Research, J.~Dy and A.~Krause, Eds., vol.~80.\hskip 1em plus 0.5em minus 0.4em\relax PMLR, 10--15 Jul 2018, pp. 4393--4402.

\bibitem{b44}
R.~Li, J.~Zhong, W.~Hu, Q.~Dai, C.~Wang, W.~Wang, and X.~Li, ``Adaptive class augmented prototype network for few-shot relation extraction,'' \emph{Neural Networks}, vol. 169, pp. 134--142, 2024.

\bibitem{b45}
A.~Krizhevsky, I.~Sutskever, and G.~E. Hinton, ``Imagenet classification with deep convolutional neural networks,'' in \emph{Advances in neural information processing systems}, 2012, pp. 1097--1105.

\bibitem{b46}
K.~He, X.~Zhang, S.~Ren, and J.~Sun, ``Deep residual learning for image recognition,'' in \emph{Proceedings of the IEEE conference on computer vision and pattern recognition}, 2016, pp. 770--778.

\bibitem{b47}
I.~O. Sigirci, H.~Ozgur, and G.~Bilgin, ``Feature extraction with bidirectional encoder representations from transformers in hyperspectral images,'' in \emph{2020 28th Signal Processing and Communications Applications Conference (SIU)}.\hskip 1em plus 0.5em minus 0.4em\relax IEEE, 2020, pp. 1--4.

\bibitem{b48}
C.~Jia and W.~He, ``Enhancerpred: a predictor for discovering enhancers based on the combination and selection of multiple features,'' \emph{Scientific reports}, vol.~6, no.~1, pp. 1--7, 2016.

\bibitem{b50}
D.~Theckedath and R.~Sedamkar, ``Detecting affect states using vgg16, resnet50 and se-resnet50 networks,'' \emph{SN Computer Science}, vol.~1, no.~2, pp. 1--7, 2020.

\bibitem{b16}
N.~Carlini and D.~Wagner, ``Towards evaluating the robustness of neural networks,'' pp. 39--57, 2017.

\bibitem{b53}
P.~Angelov and E.~Soares, ``Towards explainable deep neural networks (xdnn),'' \emph{Neural Networks}, vol. 130, pp. 185--194, 2020.

\bibitem{b54}
P.~Angelov and X.~Gu, \emph{Empirical Approach to Machine Learning}.\hskip 1em plus 0.5em minus 0.4em\relax Springer, 2019.

\bibitem{b52}
A.~Krizhevsky, ``Learning multiple layers of features from tiny images,'' pp. 32--33, 2009.

\end{thebibliography}

\end{document}